\documentclass[11pt]{article}

\usepackage[final]{acl}

\usepackage{times}
\usepackage{latexsym}
\usepackage{amsmath}
\usepackage[T1]{fontenc}
\usepackage{svg}
\usepackage{placeins} 
\usepackage{stfloats}
\usepackage[utf8]{inputenc}
\usepackage{booktabs}
\usepackage{multirow}
\usepackage{float}
\usepackage{microtype}
\usepackage{subcaption}
\usepackage{inconsolata}
\usepackage{graphicx}
\usepackage{geometry}
\usepackage[most]{tcolorbox}

\usepackage{listings}
\usepackage{longtable}
\usepackage{xcolor}

\newtcolorbox{promptbox}[1][]{
  colback=gray!5!white,   
  colframe=gray!75!black, 
  fonttitle=\bfseries,    
  title={Prompt},         
  breakable,             
  enhanced,               
  before skip=10pt,       
  after skip=10pt,        
  #1                      
}

\lstdefinelanguage{json}{
    basicstyle=\ttfamily\footnotesize, 
    breaklines=true,                  
    frame=none,                        
    keepspaces=true,                  
    showstringspaces=false,           
    keywords={true,false,null},
    keywordstyle=\color{blue}\bfseries,
    stringstyle=\color{red!70!black},
    commentstyle=\color{green!50!black},
}
\title{GRASP: Graph-Reasoning Aided Survey Planning for High-Fidelity Related Work Generation}

\author{Haoming Li \and Jessica Ouyang \\
        Department of Computer Science \\
        University of Texas at Dallas          \\
        Richardson, TX 75080                   \\
        Haoming.Li@UTDallas.edu,                  \\
        Jessica.Ouyang@UTDallas.edu }

\begin{document}
\maketitle
\begin{abstract}
Writing a literature review requires a deep understanding of the relationships among cited papers: how they build on, challenge, or offer alternative perspectives to one another. We present Graph-Reasoning Aided Survey Planning (GRASP), a framework combining LLM planning for related work generation with graph algorithms to extract key relationships among cited papers. Our two-layer graph structure consists of a Graph of Thoughts and an Argument-Counterargument Planning Network, representing the cited papers at different levels of granularity, and we apply topology-aware pruning via a Steiner tree to identify the core inter-paper relationships captured in our graph. Our citation analysis-based evaluation shows that GRASP generates RWS that closely match human-written targets in terms of the discourse roles, intents, and grouping of citations.
\end{abstract}

\section{Introduction}

Conducting a literature review is an important early step in the process of scientific research. Scientists need a deep understanding of existing work in their field, as well as the relationships among these prior works, in order to identify gaps, limitations, and challenges that their own research can address. The importance of the literature review is such that a dedicated related work section  is a key component of scientific articles, and many doctoral programs require a formal literature review as part of their candidacy qualifying exams \citep{knopf2006doing}.

The work of keeping up with an ever-growing body of existing research and drawing connections among published articles requires significant time and cognitive effort from researchers. The number of scientific publications grows exponentially \citep{bornmann2021growth}, with the result that most papers about a given topic are relatively recent \cite{wang2021science}. There are more publications, which are growing more specialized, full of jargon, and difficult to read \citep{plaven2017readability}, creating a need for tools to assist researchers in contextualizing these papers.

The success of Large Language Models in single-document summarization has fueled excitement about their potential to automate complex academic writing. However, proficiency in processing individual, self-contained documents often fails when the challenge shifts to summarizing a collection of interconnected documents, such as synthesizing related papers into a coherent literature review. 

Writing a good literature review or related work section (RWS) takes more than simply listing cited paper summaries. It requires a deeper understanding of the intricate web of relationships among papers: how they build on, challenge, or offer alternative perspectives to one another. For current models, the volume of text from multiple papers often exceeds practical context limits, making it difficult to maintain a holistic view and accurately capture inter-paper relationships \citep{li-ouyang-2024-related, liu-etal-2025-select}. Yet such relationships are precisely what human readers value in a good RWS \citep{martin2024shallow, li-ouyang-2025-explaining}. 

In this work, we propose Graph-Reasoning Aided Survey Planning (GRASP), a framework that combines LLM planning methods for related work generation with graph algorithms to extract key relationships among the cited papers. Our main contributions are as follows:
\begin{itemize}
    \item We propose a two-layer graph structure, consisting of a Graph of Thoughts \citep{besta2024graph} and an Argument-Counterargument Planning Network \citep{hua-etal-2019-argument-generation}, to represent the content of and relationships among cited papers at different levels of granularity.
    \item We introduce \textit{consensus node} merging and topology-aware pruning to identify core relationships captured in our graph, enabling the generation of a concise, high-quality RWS explicitly focused on inter-paper relationships.
    \item We conduct a \textit{citation analysis}-based evaluation that applies citation discourse and intent labeling \cite{li-etal-2022-corwa, lauscher-etal-2022-multicite}, as well as citation cluster evaluation, to demonstrate that our generated RWS closely match human-written targets in the relative importance and salient aspects of cited papers.
\end{itemize}
Our code is available at \url{https://github.com/LVenum/related_work_generation}.

\section{Related Work}

\subsection{Relationship-Focused Related Work Generation}

Early extractive RWS generation approaches \citep[][inter alia]{hoang-kan-2010-towards, hu-wan-2014-automatic} were not capable of expressing any relationships that the cited papers themselves did not explicitly discuss; early abstractive works \citep[][inter alia]{xing-etal-2020-automatic, luu-etal-2021-explaining} focused on generating citations of individual cited papers in isolation due to the model length restrictions of the time. 

More recently, \citet{chen2021capturing, chen2022target} used a graph of cited papers connected through shared keyword nodes to compute relation-aware cited paper encodings. Similarly, \citet{wang-etal-2022-multi} used a scientific information extraction system to build a knowledge graph of cited paper entities (e.g. tasks, metrics, etc.) and generated literature reviews based on lists of salient entities and relations. Like our proposed approach, these works made use of a graph structure to represent cited papers, but the graphs were used indirectly via document encodings or lists of keywords; unlike our approach, they did not use the graph structure to directly guide generation.

\citet{liu-etal-2023-causal} introduced a Causal Intervention Module (CaM) for related work generation that focused on adjusting the generation probabilities of transition words that express cited paper relationships at the beginnings of sentences (e.g. ``Furthermore, \ldots"). However, their assumption that all relevant cited paper relationships are explicitly described using such sentence-initial transition words is unrealistic.

Finally, \citet{li-ouyang-2025-explaining} extracted features for a cited paper by summarizing its edges in a citation network, while \citet{liu-etal-2025-select} walked a (co-)citation network by iteratively selecting cited paper sections to read and summarize, either continuing to the next section of a given paper or transitioning to a neighboring paper. These works, like most citation network-based approaches, can capture only relationships between papers that cite or are co-cited with each other, and thus suffer from sparsity and cold-start problems.


\subsection{Planning-Guided Survey Generation}

Planning-guided approaches break away from traditional single-shot generation and reframe the process as plan-then-realize. 
\citet{lai2024instruct} prompted an LLM to sequentially generate each section of a full survey article sequentially step-by-step, including section abstracts and subsection headers. \citet{wang2024autosurvey} built a pipeline of outline generation, subsection drafting, integration, and refinement, while \citet{yan-etal-2025-surveyforge} likewise proposed a pipeline consisting of outline generation, subsection drafting, and refinement via a final editing pass over the concatenated subsections. 

None of these works explicitly focused on the relationships among papers, relying solely on supervision from the training targets; while the target surveys should express the relationships among cited papers, they are much longer than RWS, and thus likely to contain disproportionately more specific summary content for individual papers.



\section{Methodology}

\begin{figure*}[t]
    \centering
    \includegraphics[width=\textwidth]{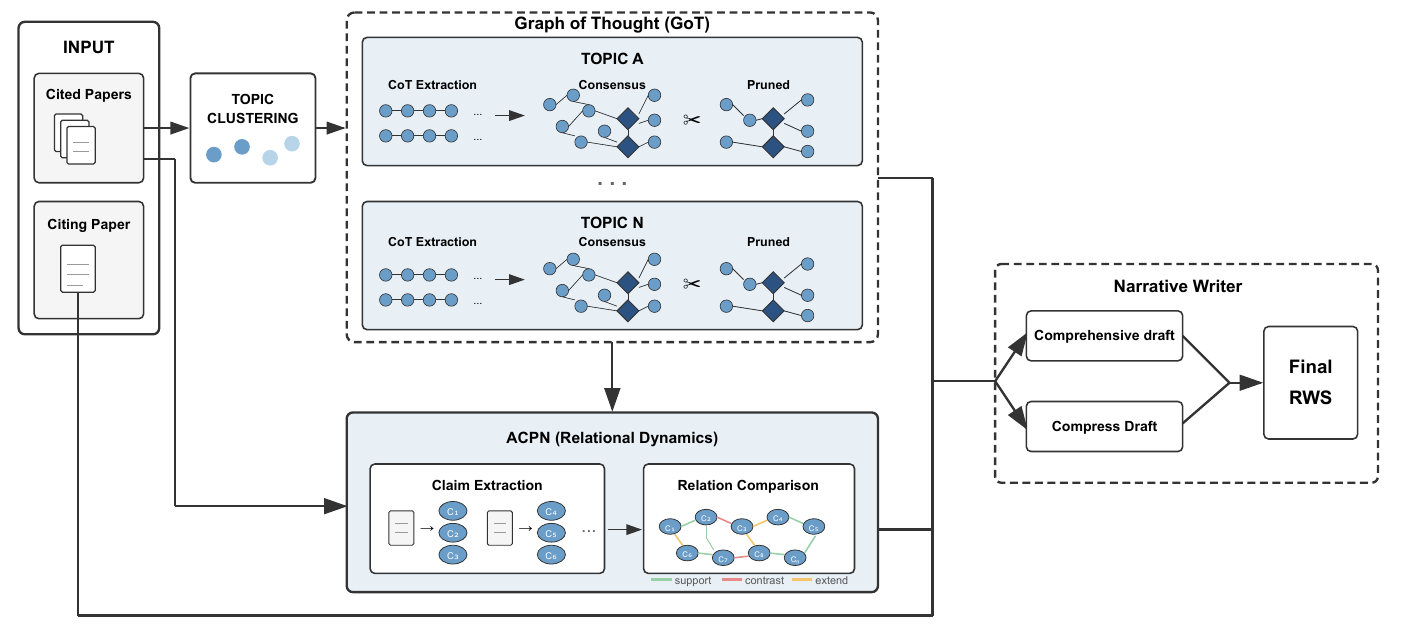} 
    \caption{Our proposed framework. The cited papers are partitioned by topic, a Chain of Thoughts is extracted for each paper, and similar thoughts are merged to form a Graph of Thoughts (GoT), which is pruned to remove peripheral thoughts. Claims are extracted from each paper and, along with the paper's GoT nodes, are used to classify the relationships between pairs of cited papers, producing the Argument-Counterargument Planning Network (ACPN). Finally, the Writer module drafts, compresses, and refines the output related work section (RWS).}
    \label{fig:my_framework}
\end{figure*}

Given a citing paper $P$ (excluding the target RWS) and a set of cited papers $R = \{ R_1,R_2,\ldots, R_m \}$, our goal is to generate the RWS of $P$ using $R$. Figure \ref{fig:my_framework} shows an overview of our GRASP framework, and Appendix \ref{app:prompts} shows our LLM prompts.

\subsection{Graph of Thoughts Layer}
We combine the semantic reasoning capabilities of Large Language Models (LLMs) with structural guidance via a Graph of Thoughts \citep[GoT;][]{besta2024graph}. While LLMs excel at extracting local information, they struggle with macro-level structure and thematic coherence. We address this problem by using traditional graph algorithms to prune the GoT, retaining only the structural backbone and most central content.

\subsubsection{Topic-Partitioned Graph Construction}
A RWS typically covers distinct thematic clusters\footnote{We empirically validate this assumption in Appendix \ref{app:topics}.}, so we partition the set of cited papers $\mathcal{R}$ into $k$ topic-specific subsets, $\mathcal{R} = \bigcup_{i \in [1, k]} T_i$, using an LLM-prompting clustering approach. To ensure scalability and coherence, we construct a local reasoning graph $G_i$ for each topic $T_i$ independently. 

\paragraph{Sequential Node Extraction}
Within each topic $T_i$, we employ Chain-of-Thought prompting \cite{wei2022chain} to extract the logical workflow of each cited paper $r_j \in T_i$. We prioritize key sections that are most likely to discuss a paper's unique contributions --- Abstract, Introduction, and Conclusion --- by limiting the other sections to at most one single thought each, ensuring that the majority of extracted thoughts capture core claims rather than peripheral details. These thoughts form the \textit{sequential nodes} of our graph, linked by directed edges to preserve the intra-paper narrative flow.

\paragraph{Consensus Node Formation}
To explicitly capture inter-paper relationships, we use an LLM prompt to assess the semantic similarity between pairs of sequential nodes in $T_i$. When nodes from different papers exhibit significant semantic overlap, they are merged into a \textit{consensus node} that summarizes the original sequential nodes. 

\begin{figure*}[t]
    \centering
    \includegraphics[width=\textwidth]{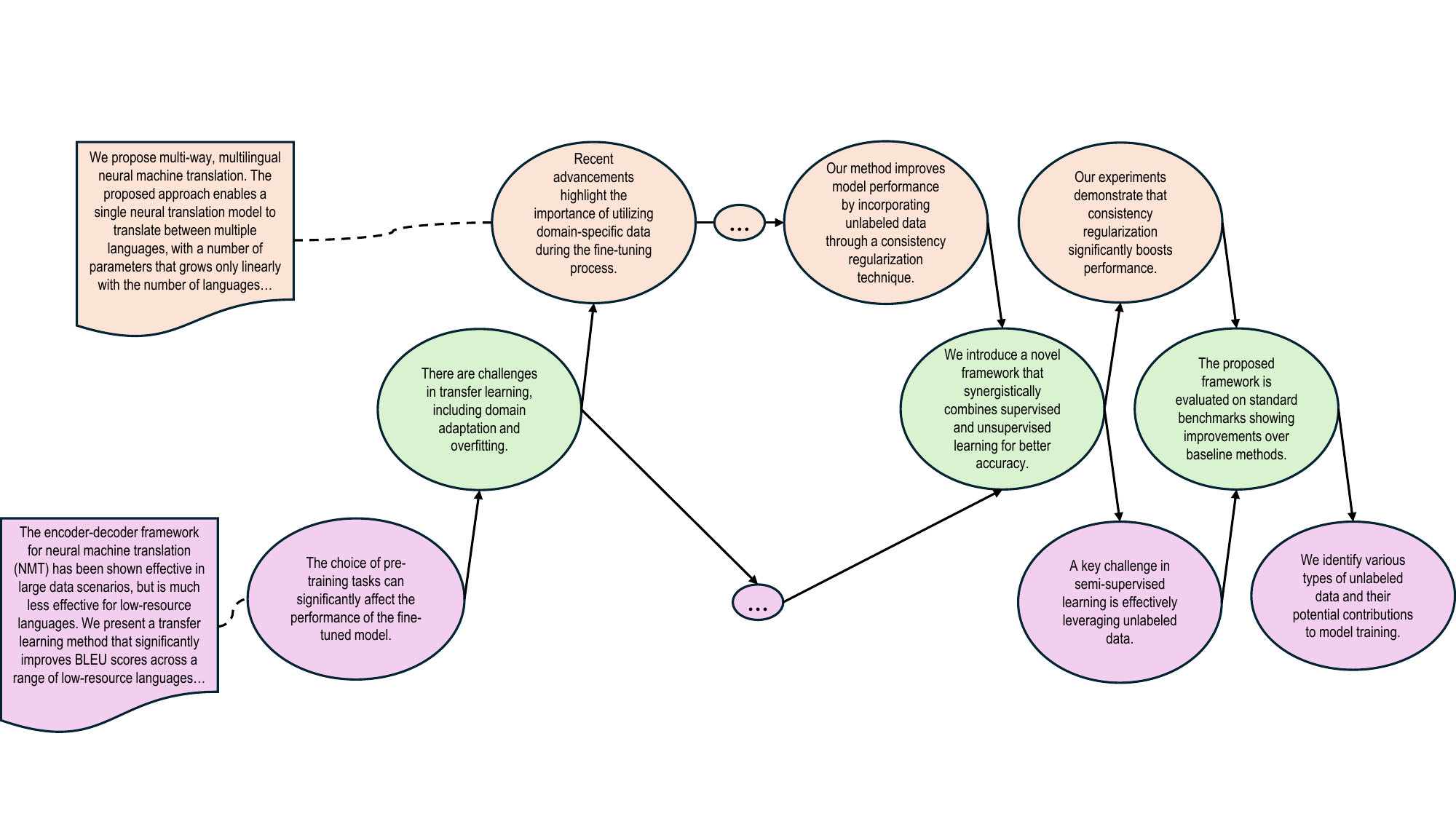} 
    \caption{An excerpt from a Graph of Thoughts containing ``Multi-Way, Multilingual Neural Machine Translation with a Shared Attention Mechanism" \citep[][top, shown in orange]{firat-etal-2016-multi} and ``Transfer Learning for Low-Resource Neural Machine Translation" \citep[][bottom, shown in pink]{zoph-etal-2016-transfer}. Shared \textit{consensus nodes} are shown in green.}
    \label{fig:consensus}
\end{figure*}

Figure \ref{fig:consensus} shows example sequential and consensus nodes in an abbreviated GoT.

\subsubsection{Topology-Aware Pruning}
The raw reasoning graphs are noisy, containing fine-grained details specific to individual papers that are excessive for a high-level RWS. We prune each topic subgraph $G_i=(V_i, E_i)$ to identify and retain only a semantic backbone of nodes that are salient to multiple cited papers. We formulate this pruning step as a \textbf{Steiner Tree} problem, which seeks the minimum-weight subgraph that connects a subset of critical \textbf{terminal} nodes.

\paragraph{Terminal Identification}
We identify the set of terminal nodes $S \subset V_i$ that represent the main inter-paper relationships in $G_i$:
\begin{itemize}
    \item \textbf{Consensus nodes:} These nodes aggregate information shared among multiple papers, representing pivotal points of comparison.
    \item \textbf{High-centrality sequential nodes:} Consensus nodes are relatively rare (averaging 1.79 per graph), so we calculate the \textit{betweenness centrality} \cite{freeman1977set} for all sequential nodes and retain the top $20\%$ as additional terminals\footnote{We tune this betweenness centrality threshold in Appendix \ref{app:centrality}}. We hypothesize that high-centrality nodes are topologically close to consensus nodes and serve as connective tissue within the graph, capturing shared methodological contexts common to multiple works.
\end{itemize}

\paragraph{Steiner Tree Approximation}
Since the Steiner tree problem is NP-hard, we implement a 2-approximation algorithm \cite{wu2004spanning}. First, we construct a metric closure graph, where the edge weight between two terminals $u, v \in S$ is their shortest path distance in the original graph $G_i$. We then compute the minimum spanning tree (MST) of this closure and map the edges of the MST back to their corresponding paths $G_i$ to form the pruned subgraph $G'_i$.

The Steiner tree rigorously removes dead ends and redundant paths that do not contribute to the connectivity of the cited papers' shared ideas, as captured by the terminals. This pruning step ensures that our Writer module receives a concise, topologically streamlined blueprint that links all main ideas with a minimum of irrelevant, paper-specific details.

\subsection{Argument-Counterargument Planning Network Layer}
The GoT captures fine-grained similarities among cited papers but does not explicitly model the global relationships between papers. Two papers that do not share exact approaches in common may still present supporting evidence for a shared hypothesis, offer contrasting findings, or address difference aspects of the same problem. Recognizing these relationships is essential for generating RWS that accurately characterize the research landscape.

We use an Argument-Counterargument Planning Network \citep[ACPN;][]{hua-etal-2019-argument-generation} to construct a paper-level argumentative relation graph over the set of cited papers $R = \{ R_1, R_2, ..., R_m \}$: the ACPN produces a directed graph $H$ where each edge $(R_i, R_j)$ is labeled with a relation $r \in \{ \text{support}, \text{contrast}, \text{neutral} \}$.

First, we extract the core claims from each cited paper $R_i$ using LLM prompting. Unlike the fine-grained, section-specific summaries in the GoT described in the previous section, these claims capture the overarching theses of each paper. Then, representing each paper by its claims and GoT nodes, we classify the pairwise relationships between papers. 

The resulting ACPN graph provides explicit signals for generation. Edges labeled as \textit{contrast} indicate opportunities to discuss methodological differences or conflicting results, while \textit{support} edges identify papers that can be grouped to establish consensus or trace the development of an idea.

\subsection{RWS Writer}

The final component is an LLM-based Writer module that uses citing paper context and the structural insights of the GoT and ACPN to generate a coherent RWS. To preserve graph topological information, we serialize both graph layers into a structured JSON node-link format that can be inserted into our Writer prompt, allowing the LLM to traverse the GoT and ACPN paths. To balance the competing requirements of technical completeness and linguistic conciseness, we implement a three-stage drafting strategy within our prompt design:
\begin{enumerate}
    \item \textbf{Comprehensive Draft:} The model first generates a verbose RWS draft explicitly incorporating every node from the graph to ensure maximal information coverage.
    \item \textbf{Semantic Compression:} The model then produces a compressed version of RWS, removing redundancy and verbosity from the first draft while optimizing for information density and logical progression.
    \item \textbf{Final Merging:} Finally, the model performs a reconciliation step that fuses the structural fidelity of the Comprehensive draft with the brevity of the Compressed version. We explicitly instruct guide the Writer to remove duplicate citations and smooth out transitions, producing a more polished and fluent RWS (see Section \ref{sec:ablation}).
\end{enumerate}

We use a post-processing step to standardize the format of citation markers for evaluation and to correct hallucinated author names or years, which we describe in Appendix \ref{app:citation-management}.

\section{Experimental Settings}

We use GPT-4o-mini\footnote{https://platform.openai.com/docs/models/gpt-4o-mini} and as our base model because of its competitive balance of performance and cost. 
Appendix \ref{app:prompts} shows our prompts, and Appendix \ref{app:cost} gives the run time and cost of our experiments.

\subsection{Dataset}
To evaluate our proposed framework, we use the OARelatedWork test set \cite{docekal2024oarelatedworklargescaledatasetrelated}, which consists of 1,878 papers from the CORE \cite{Knoth2023COREAG} and S2ORC \cite{lo-etal-2020-s2orc} academic writing datasets. Unlike other collections of scientific articles, the papers in OARelatedWork are curated to ensure the availability of their cited papers: for each target RWS, the full text of at least one cited paper from each citation span is included in the dataset. 
We use a subset of 1,350 papers from the test set; we filter out target RWS that contain fewer than three cited papers in order to focus our evaluation on capturing relationships among cited papers. 

Additionally, we clean the target RWS by removing references to cited papers that are not included in OARelatedWork, as only one paper per citation span is guaranteed to be available. Since it is impossible for our approach (or the baselines) to write about papers that are not included in its input, we remove any citation marks whose papers are not available; we further delete any sentence whose citation marks have all been removed in this way\footnote{Such sentences are not supposed to exist based on the construction of OARelatedWork, but in practice we found that almost 18\% of sentences containing citation marks were missing all of their cited papers.}.


\subsection{Baselines}

We compare our proposed approach with two recent graph-based RWS generation works:

\vspace{-.5\baselineskip}

\paragraph{\citet{li-ouyang-2025-explaining}} (hereafter L\&O for brevity) uses a citation network to extract relationship features for each cited paper by summarizing its incoming and outgoing edges, capturing both how other papers cite it and what it says about other papers. We use their provided LLM prompts to generate RWS for evaluation.

\paragraph{Select, Read, Write \citep[SRW;][]{liu-etal-2025-select}} uses a multi-agent framework with a Select module guided by a (co-)citation network, allowing it to transition to an adjacent paper from the one it was previously reading. As with L\&O, we use the provided LLM prompts to generate RWS for evaluation.

\vspace{.5\baselineskip}

We also compare against two graphless baselines:

\vspace{-.5\baselineskip}

\paragraph{No-graph} uses the same topic clusters, extracted thoughts and claims, and RWS Writer prompt as GRASP, but simply concatenates all of the thoughts and claims sequentially, instead of building (and pruning) the GoT and ACPN. This baseline captures the contribution of our GoT and ACPN's \textit{structure}, as opposed to their information content.

\paragraph{Direct} generation uses GPT-5-mini\footnote{https://developers.openai.com/api/docs/models/gpt-5-mini} to implement a naive, strong LLM baseline that generates RWS from the concatenation of cited paper texts. This baseline captures the capability of a strong, long context model to directly generate RWS with no explicit guidance from the thoughts, claims, and graph structures modeled in GRASP. 

\vspace{.5\baselineskip}

Appendix \ref{app:generated-examples} shows an example (cleaned) target RWS from OARelatedWork and the corresponding generated RWS from GRASP and the baselines.

\subsection{Citation Analysis-Based Metrics}

Evaluating the quality of RWS is challenging. On the one hand, traditional summarization metrics like ROUGE and BERTScore, as well as generic LLM-as-judge evaluations, demonstrate poor correlation with human judgments on scientific texts \cite{chen2024rethinking}. On the other hand, human evaluation requires recruiting judges with a strong familiarity with the set of cited papers, which is difficult to achieve for an open-domain dataset like OARelatedWork.

We adopt an automatic, \textit{citation analysis}-based evaluation consisting of four facets.

\paragraph{Sentence Discourse Roles} Following our L\&O baseline, we use the CORWA tagger \cite{li-etal-2022-corwa} to label each sentence in an RWS with a discourse role: \textit{single} or \textit{multi}-paper summary citation, \textit{narrative} citation, \textit{reflection} on the citing paper, or introduction/\textit{transition} sentence. Because there may not be an exact mapping between sentences, and the total number of sentences in the generated and target RWS varies, we calculate the \textbf{ratio difference} of each discourse role, evaluating the balance of paper-specific details (single paper summaries) and inter-paper relationships (multi-paper summaries, narrative citations, and reflections) in the generated RWS.

\paragraph{Citation Importance} The CORWA tagger additionally labels each citation as \textit{dominant} if it is the main focus of its sentence, or \textit{reference} if it is peripheral, roughly corresponding to the level of detail it receives in the RWS. We calculate accuracy and F1 score for each cited paper, treating its dominant/reference status in the target RWS as its ground truth class. This facet evaluates the generated RWS on expressing the relative importance of each cited paper. 

\begin{table*}[ht]
  \centering
  \small 
  \resizebox{\textwidth}{!}{
    \begin{tabular}{l ccc ccc ccc ccc cc cc}
    \toprule
    & \multicolumn{3}{c}{\textbf{ROUGE-1}} & \multicolumn{3}{c}{\textbf{ROUGE-2}} & \multicolumn{3}{c}{\textbf{ROUGE-L}} & \multicolumn{3}{c}{\textbf{BERTScore}} & \multicolumn{2}{c}{\textbf{BLEU}} & \multicolumn{2}{c}{\textbf{METEOR}} \\
    \cmidrule(lr){2-4} \cmidrule(lr){5-7} \cmidrule(lr){8-10} \cmidrule(lr){11-13} \cmidrule(lr){14-15} \cmidrule(lr){16-17}
    \textbf{Model}  & P     & R     & F1    & P     & R     & F1    & P     & R     & F1    & P     & R     & F1    & Macro & Micro & Macro & Micro \\
    \midrule
    SRW   & 0.489 & 0.672 & 0.529 & 0.341 & 0.522 & 0.388 & 0.394 & 0.577 & 0.439 & 0.297 & 0.487 & 0.389 & 31.86 & 37.59 & 0.458 & 0.410 \\
    L\&O & 0.517 & 0.982 & 0.664 & 0.507 & 0.975 & 0.655 & 0.510 & 0.977 & 0.658 & 0.715 & 0.926 & 0.817 & 51.12 & \underline{54.67} & 0.680 & 0.482 \\
    \midrule
    Direct & 0.377 & 0.984 & 0.532 & 0.369 & 0.975 & 0.523 & 0.371 & 0.977 & 0.526 & 0.662 & 0.920 & 0.785 & 37.91 & 40.57 & 0.600 & 0.446 \\
    No-graph & 0.620 & 0.979 & 0.747 & 0.611 & 0.974 & 0.740 & 0.615 & 0.975 & 0.743 & 0.726 & 0.924 & 0.822 & 60.85 & 64.54 & 0.764 & 0.498 \\
    \midrule
    GRASP (unpruned) & 0.631 & 0.977 & \underline{0.751} & 0.622 & 0.973 & \underline{0.745} & 0.626 & 0.975 & \underline{0.748} & 0.731 & 0.934 & \underline{0.829} & \underline{61.36} & 48.75 & \underline{0.768} & \underline{0.488} \\
    GRASP (pruned) & 0.653 & 0.978 & \textbf{0.771} & 0.644 & 0.974 & \textbf{0.766} & 0.648 & 0.975 & \textbf{0.769} & 0.738 & 0.932 & \textbf{0.832} & \textbf{63.42} & \textbf{68.09} & \textbf{0.778} & \textbf{0.509} \\
    \bottomrule
    \end{tabular}%
  }
  \caption{Evaluation with traditional text generation metrics. We report average ROUGE, BERTScore, BLEU, and METEOR. Best results are highlighted in \textbf{bold}, with the runner-up \underline{underlined}.}
  \label{tab:main_results}%
\end{table*}%

\paragraph{Citation Intent} We use the MultiCite tagger \cite{lauscher-etal-2022-multicite} to label the function of each citation as giving \textit{background} information, \textit{motivating} the citing paper, \textit{using} or \textit{extending} a given approach, comparing \textit{similarities} or \textit{differences} among papers, or suggesting \textit{future work}. We again calculate accuracy and F1 score for each cited paper, treating its intent in the target RWS as its ground truth class. This facet evaluates the generated RWS on the capturing most salient aspect of each cited paper. 

\paragraph{Citation Co-Occurrence} Our SRW baseline \cite{liu-etal-2025-select} evaluates relationships among cited papers via a graph where edges between papers represent citation within the same sentence. We argue that this dependence on intra-sentence co-occurrence is too strict; Liu et al. find only two to three edges per RWS on average. Further, their metrics of edge count, node degree, and clustering coefficient consider only the \textit{frequency} of citation co-occurrence, but not \textit{which} papers are cited together. 

We instead evaluate citation co-occurrence at the paragraph level, with edges between papers cited within the same paragraph. We use \textbf{edge-connected Jaccard similarity} to measure grouping fidelity by comparing each cited paper's neighbors in a generated RWS paragraph to those in its target RWS paragraph. Additionally, we evaluate the relative \textit{ordering} of citations within a generated RWS (e.g. how foundational works should be discussed before papers that extend their approaches) by enumerating its citation marks in order of occurrence and calculating \textbf{Kendall's $\tau$} with the target RWS ordering.  

\section{Results and Discussion}
\label{sec:results}
\subsection{Text Generation Metrics}

Table \ref{tab:main_results} evaluates the performance of our proposed GRASP framework using traditional text generation metrics: ROUGE-1, -2, and -L; BERTScore; BLEU; and METEOR. GRASP outperforms both baselines across all metrics. 

In terms of lexical overlap-based metrics, our approach captures significantly more relevant $n$-grams and maintains better structural coherence than the baselines. While L\&O achieves high ROUGE recall, their corresponding precision is notably lower, suggesting their approach may be over-generating content to maximize coverage at the expense of conciseness (this interpretation is further supported by Appendix Table \ref{tab:generationstats}). In contrast, GRASP maintains comparable recall while delivering substantially higher precision, resulting in a higher-quality RWS. Beyond lexical overlap, GRASP also demonstrates superior semantic alignment with the target RWS, as evidenced by its lead in BERTScore. 

\begin{figure}[t]
    \centering
        \includegraphics[width=\linewidth]{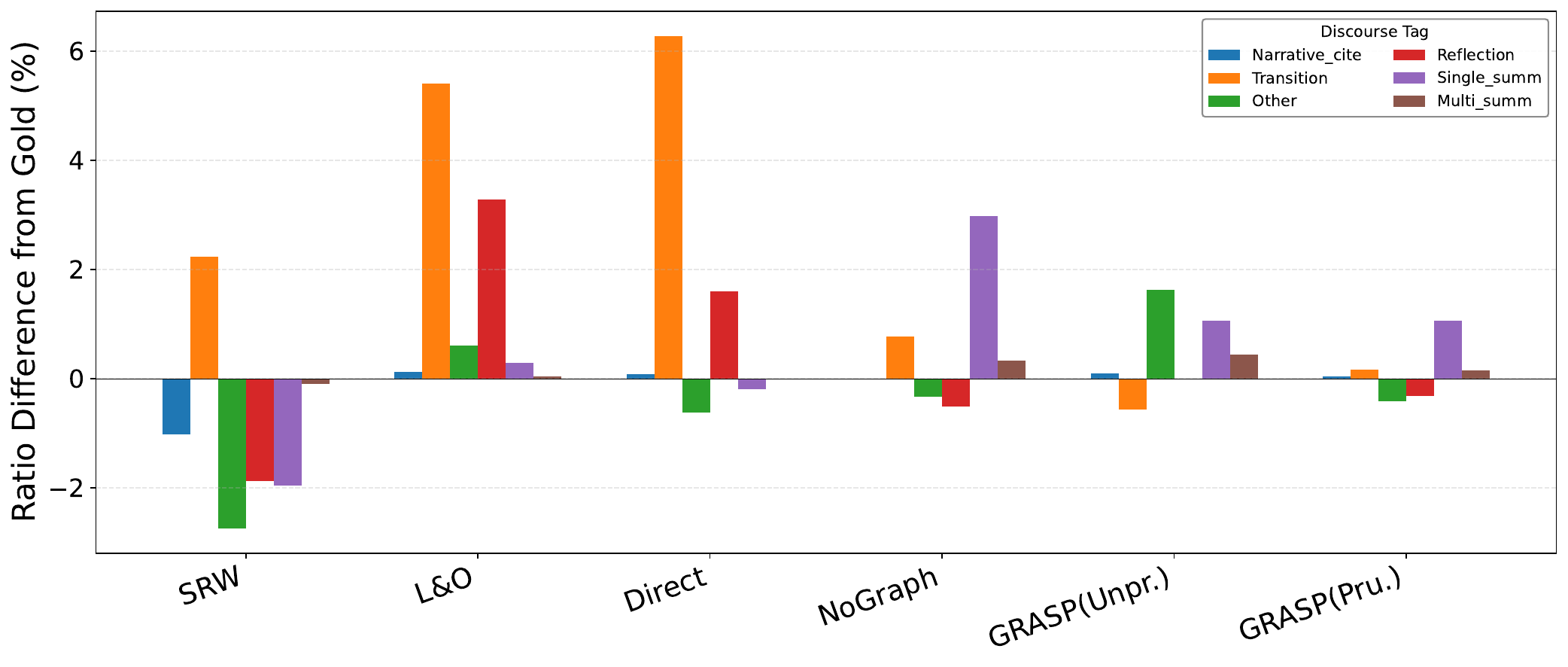}
        \caption{Ratio difference of discourse roles by method; our proposed approaches match the target discourse distribution more closely (shorter bars) than the baselines.}
        \label{fig:ratio_diff_method}




    
\end{figure}

\begin{table*}[ht]
  \centering
  \label{tab:discourse_breakdown}
  \resizebox{\textwidth}{!}{%
    \begin{tabular}{l ccc ccc ccc ccc ccc ccc ccc}
    \toprule
    \multirow{2}{*}{\textbf{Model}} & \multicolumn{3}{c}{\textbf{Background}} & \multicolumn{3}{c}{\textbf{Differences}} & \multicolumn{3}{c}{\textbf{Extends}} & \multicolumn{3}{c}{\textbf{Future Work}} & \multicolumn{3}{c}{\textbf{Motivation}} & \multicolumn{3}{c}{\textbf{Similarities}} & \multicolumn{3}{c}{\textbf{Uses}} \\
    \cmidrule(lr){2-4} \cmidrule(lr){5-7} \cmidrule(lr){8-10} \cmidrule(lr){11-13} \cmidrule(lr){14-16} \cmidrule(lr){17-19} \cmidrule(lr){20-22}
     & P & R & F1 & P & R & F1 & P & R & F1 & P & R & F1 & P & R & F1 & P & R & F1 & P & R & F1 \\
    \midrule
    SRW & 0.163 & 0.518 & 0.248 & 0.313 & 0.464 & 0.374 & 0.125 & 0.700 & 0.212 & 0.000 & 0.000 & 0.000 & 0.091 & 0.581 & 0.158 & 0.027 & 0.211 & 0.048 & 0.489 & 0.319 & 0.387 \\
    L\&O & 0.670 & 0.965 & 0.791 & 0.590 & 0.960 & 0.731 & 0.340 & 1.000 & 0.508 & 0.429 & 1.000 & 0.600 & 0.274 & 0.983 & 0.429 & 0.492 & 0.955 & 0.650 & 0.827 & 0.950 & 0.884 \\
    \midrule
    Direct & 0.320 & 0.964 & 0.481 & 0.090 & 0.957 & 0.481 & 0.054 & 1.000 & 0.103 & 0.400 & 1.000 & 0.571 & 0.205 & 0.988 & 0.339 & 0.343 & 0.966 & 0.505 & 0.352 & 0.983 & 0.519\\
    No-graph & 0.649 & 0.935 & 0.766 & 0.0.838 & 0.954 & 0.892 & 0.962 & 1.000 & \textbf{0.981} & 0.750 & 1.000 & 0.857 & 0.536 & 0.974 & 0.691 & 0.977 & 0.955 & \textbf{0.966} & 0.974 & 0.943 & 0.958\\
    \midrule
    GRASP (unpruned) & 0.700 & 0.965 & \underline{0.811} & 0.880 & 0.960 & \textbf{0.918} & 0.940 &1.000 & \underline{0.971} & 0.750 & 1.000 & \underline{0.857} & 0.635 & 0.983 & \underline{0.772} & 0.977 & 0.955 & \textbf{0.966} & 0.985& 0.950 & \underline{0.967} \\
    GRASP (pruned) & 0.728 & 0.964 & \textbf{0.829} & 0.844 & 0.960 & \underline{0.898} & 0.862 & 0.980 & 0.917 & 1.000 & 1.000 & \textbf{1.000} & 0.684 & 0.983 & \textbf{0.806} & 0.955 & 0.955 & \underline{0.955} & 0.991 & 0.950 & \textbf{0.970} \\
    \midrule
    \bottomrule
    \end{tabular}%
  }
  \caption{Citation intent fidelity. We report average Precision, Recall, and F1-score, measured against the intent of each cited paper in the target RWS. Best results are highlighted in \textbf{bold}, with the runner-up \underline{underlined}.}
  \label{tab:multicite}
\end{table*}

\subsection{Sentence Discourse Roles}

Fig \ref{fig:ratio_diff_method} compares the distribution of sentence discourse roles\footnote{The \textit{other} role (shown in green in Figure \ref{fig:ratio_diff_method}) is a catch-all category for unexpected sentence formats, usually subsection or paragraph headers.} in generated RWS. Both baselines exhibit significant deviations from the target distribution. L\&O disproportionately over-generates \textit{introduction/transition} and \textit{reflection} sentences (shown in orange and red), suggesting a tendency make generalized statements and focus on the citing paper, rather than providing enough information about the cited papers (see also Appendix Table \ref{tab:generationstats}). SRW likewise over-generates \textit{introduction/transition} while under-generating other sentence types. In contrast, our Steiner-pruned GRASP achieves a distribution that aligns closely with the target, capturing a natural balance of summarization, transition, and synthesis.

The discourse evaluation also demonstrates the importance of our GoT and ACPN graph structures. We see that the No-graph baseline, while using the same thoughts and claims as GRASP, devolves into a list of individual paper summaries. The deviation for Single Summarization sentences nearly triples, and the No-graph baseline under-generates complex structural sentence types like Transition and Reflection. We can further see this difference reflected in the citation density of the generated RWS: the No-graph baseline significantly increases the average citation spans per sentence by +28.38\% compared to the ground truth, contrasted with the much smaller density difference of only -7.37\% for pruned GRASP.

\begin{table}[t]
  \centering
    \resizebox{\linewidth}{!}{%
    \begin{tabular}{l ccc ccc}
    \toprule
    & \multicolumn{3}{c}{\textbf{Dominant}} & \multicolumn{3}{c}{\textbf{Reference}} \\
    \cmidrule(lr){2-4} \cmidrule(lr){5-7}
    \textbf{Model} & P & R & F1 & P & R & F1 \\
    \midrule
    SRW   & 0.263 & 0.263 & 0.263 & 0.389 & 0.523 & 0.446 \\
    L\&O & 0.710  & 0.928 & 0.804 & 0.944 & 0.973 & \textbf{0.958} \\
    \midrule
    Direct & 0.773 & 0.921 & \underline{0.841} & 0.925 & 0.976 &  \underline{0.949}\\
    No-graph & 0.673 & 0.906 & 0.773 & 0.842 & 0.949 &  0.892\\
    \midrule
    GRASP (unpruned) & 0.675 & 0.915 & 0.777 & 0.918 & 0.983 & \underline{0.949} \\
    GRASP (pruned) & 0.916 & 0.928 & \textbf{0.922} & 0.917 & 0.983 & \underline{0.949} \\
    \bottomrule
    \end{tabular}%
    }
    \caption{Citation importance fidelity. We report average Precision, Recall, and F1-score, measured against the \textit{Dominant}/\textit{Reference} status of each cited paper in the target RWS.}
  \label{tab:domref}
\end{table}

\subsection{Citation Importance}
\label{sec:domref}

Table \ref{tab:domref} shows the citation importance fidelity of a cited paper in a generated RWS to its ground truth \textit{dominant}/\textit{reference} status in the target RWS. Pruned GRASP achieves the strongest performance on \textit{Dominant}-type citations, while only slightly underperforming L\&O on \textit{Reference}-type citations. Our approach successfully identifies the most important cited papers that are salient to the narrative backbone of the RWS and thus deserve the longer, more detailed \textit{Dominant} citations.

Appendix Table \ref{tab:generationstats} further shows that, in terms of total length, SRW is extremely concise and over-generates the shorter, more generalized \textit{Reference}-type citations. L\&O and unpruned GRASP generate over-length RWS containing disproportionately many \textit{Dominant}-type citations, indicating a tendency to generate individual paper descriptions, rather than focusing on inter-paper relationships. Pruned GRASP reduces verbosity by filtering topologically peripheral GoT nodes, thereby reducing the sentence count and producing the \textit{dominant}/\textit{reference} ratio closest to that of the target RWS. 

\subsection{Citation Intent}
Table \ref{tab:multicite} shows the citation intent fidelity of a cited paper in a generated RWS to its ground truth intent in the target. GRASP demonstrates strong performance in capturing relationships in the development of cited papers, especially \textit{Differences}, \textit{Extends}, and \textit{Similarities}. This result shows that the explicit inter-paper argumentative relation modeling in our ACPN module successfully guides the Writer module to accurately describe how each paper builds upon or diverges from prior methodologies. 

GRASP also captures the broader research context. We achieve a perfect F1 score on the relatively rare \textit{Future Work} category and outperform the baselines on \textit{Motivation}, suggesting that our consensus-based pruning strategy effectively emphasizes high-level research objectives and speculative directions discussed in the cited papers. In contrast, the baselines achieve their highest scores in the \textit{Background} and \textit{Uses} categories, focusing more on generalized problem statements and descriptions of specific approaches.

\subsection{Citation Co-Occurrence}

\begin{table}[t]
\centering
\scriptsize
\begin{tabular}{l cc}
\toprule
\textbf{Model} & \textbf{Edge Jaccard} & \textbf{Kendall's $\tau$} \\ 
\midrule
SRW & 0.569 & 0.628 \\ 
L\&O & 0.791 & \textbf{0.887} \\ 
\midrule
Direct & 0.017 & 0.267 \\
No-graph & 0.331 & 0.151 \\
\midrule
GRASP (unpruned) & \underline{0.835} & 0.885 \\ 
GRASP (pruned) & \textbf{0.847} & \underline{0.886} \\ 
\bottomrule
\end{tabular}%
\caption{Evaluation of citation co-occurrence. We report edge-connected Jaccard similarity (citation grouping) and Kendall's Tau (citation ordering).}
\label{tab:citation_structure}
\end{table}

Table \ref{tab:citation_structure} evaluates the grouping and ordering of citations in generated RWS, as compared to the target. GRASP significantly outperforms the baselines in edge-connected Jaccard similarity, measuring the the co-occurrence of cited papers at the paragraph level. This grouping fidelity is due to the explicit topic partitioning step upstream of our Graph of Thoughts construction. By partitioning the set of cited papers into topic clusters prior to subgraph generation, we enforce an overall graph topology where topically related papers form densely connected subgraphs. This structural prior acts as a blueprint for the Writer, guiding it to generate paragraphs that discuss groups of closely related papers, thereby minimizing the scattering of related papers noted by \citet{li-ouyang-2025-explaining}. 

GRASP also achieves a high Kendall's $\tau$, measuring the correlation of the generated citation ordering with that of the target RWS; our scores are comparable to L\&O, which used a simple --- and apparently very strong --- chronological ordering heuristic. This result shows that GRASP successfully balances local citation grouping with global ordering, producing an RWS that is both information-rich and coherent. 

\subsection{Effect of GoT Pruning}
A key contribution of our work is the use of Steiner Tree-based pruning to filter irrelevant nodes from the cited paper GoT. Our experimental results validate this design choice: pruned GRASP outperforms the unpruned variant across all text generation metrics (Table \ref{tab:main_results}), discourse roles (Figure \ref{fig:ratio_diff_method}), and citation importance/co-occurrence metrics (Tables \ref{tab:domref}-\ref{tab:generationstats}). As discussed in Section \ref{sec:domref}, the citation importance evaluation highlights the success of our pruning approach at removing peripheral nodes, effectively reducing noise and verbosity without discarding critical information and allowing the Writer module to focus on the semantic backbone of the cited paper set.

\subsection{Ablation Studies}
\label{sec:ablation}

We evaluate the contribution of each of our two graph layers --- the GoT capturing topical grouping and cited paper content and the ACPN capturing argumentative relations between papers --- as well as the effect of our Writer module's drafting stages. Due to space limitations, the quantitative results are found in Appendices \ref{app:ablation} and \ref{app:drafts}, and we briefly summarize them here.

\paragraph{GoT versus ACPN and Direct ACPN}

As shown in Appendix \ref{app:ablation}, the GoT-only variant maintains strong recall performance on text generation metrics and citation intent fidelity, but suffers from lowered precision, while the ACPN-only variant performs significantly worse in both precision and recall. These findings reflect the design of our two graph layers. The GoT encodes the cited paper content, so the GoT-only model is able to generate most of the target RWS content. The ACPN provides higher-level guidance on inter-paper argumentative relationships, aiding the Writer module in focusing on the most salient aspects of the cited papers, but the ACPN alone does not capture enough cited paper content to generate a good RWS. In terms of sentence discourse roles, both ablated variants over-generate single cited paper summaries, suggesting that both graph layers are needed to focus the Writer on inter-paper relationships.

We also test a simpler variant of the ACPN that uses an LLM prompt to directly classify the relationship between two papers based on their text, omitting the intermediate claim extraction step and the use of GoT node text as an additional input to the relationship classification decision. We find that, without explicit claims to anchor comparisons among cited papers, the variant’s ability to match ground truth citation intent collapses. The degradation in citation intent performance is most severe in relations that require specific evidence, such as \textit{Similarities/Differences} and \textit{Extends}. The semantic similarity to the ground truth RWS also decreases, reflecting the loss of concrete, claim-backed technical content in the generated text. Finally, lacking concrete extracted claims, the direct ACPN variant over-generates filler sentences in the form of too many \textit{Transitions} and \textit{Reflections}.

\paragraph{Compressed versus Final Drafts}
As shown in Appendix \ref{app:drafts}, the Compressed and Final drafts produced by our Writer module achieve very close performance across metrics. The Final draft trades a slight dip in lexical metrics for improved discourse role and \textit{dominant}/\textit{reference} ratios, citation intent fidelity, and citation grouping/ordering. Qualitatively, we find that the Final draft is more expressive and fluent; the Compressed draft relies on ambiguous or unexpected word choice in order to reduce the overall word count, often falling into noticeably AI-like phrasing (see Appendix Figures \ref{fig:example-pruned} and \ref{fig:example-compressed} to compare example Compressed and Final drafts).


\section{Conclusion}
We have proposed GRASP, a structure-aware framework for automated RWS generation. By combining a cited paper Graph-of-Thoughts for topical clustering and detailed relationship extraction with an Argument-Counterargument Planning Network for high-level inter-paper relationships and transition planning, our approach effectively models both cited paper content and narrative flow. Our Steiner Tree pruning step enhances the conciseness and relevance of generated RWS by explicitly focusing on consensus and high-centrality nodes. To validate our framework, we introduced several citation analysis-based metrics, including citation intent fidelity and Edge Jaccard for citation grouping. Our experiments demonstrate that GRASP significantly outperforms existing baselines, producing RWS that closely follow the structure and citation distribution of human-written targets.

\section*{Limitations}

Our current framework achieves strong performance but still has several limitations. First, we assume the set of cited papers' full texts are given as input. While this assumption is used in most existing related work generation approaches, it may not reflect a realistic use case, as a user would need to manually collect and pre-process the cited paper PDFs. 

Second, constructing the GoT and ACPN for a large number of cited papers is expensive and time-consuming; since the GoT is an input to the ACPN construction module, these two graphs cannot be built concurrently. Further, consensus nodes are relatively rare in our GoT graph. To achieve a more aggressive merging consensus nodes with high semantic overlaps (at the cost of significant computational overhead) future work could use specialized semantic similarity or entailment models for pairwise scoring across thoughts.

Third, we partition the cited papers into disjoint topic clusters, which may be overly restrictive. While we validate this design choice for the majority of cited papers in Appendix \ref{app:topics}, it is still possible for some cited papers to be relevant to more than one topic, and restricting their discussion to only one topic could weaken the overall informativeness of the generated RWS. 

Finally, our framework is not intended to replace human RWS writing, but only as an assistive tool. We do not check for plagiarism, and while we do check for hallucinated citation marks (i.e. non-existent author/year combinations), LLM-generated RWS can still contain factually incorrect statements.


\bibliography{custom}

\appendix

\section{MutuallyExclusive Topics}
\label{app:topics}

To empirically validate our assumption of mutually exclusive topics, we conduct an analysis on the ground-truth RWS in our test set. We used Sentence Transformers all-MiniLM-L6-v2 \cite{reimers-2019-sentence-bert} to cluster the human-written RWS sentences into topics across several semantic similarity thresholds, ranging from 0.1 to 0.9, and then counted how often a cited paper appeared in multiple topics within the same RWS. 

We found that, across all 8,467 cited papers in the test set ground truth RWS, cross-topic citations are rare. At a semantic similarity threshold of 0.20, which averages about 5 distinct topics per RWS, only 712 papers (8.41\%) are cited across multiple topics. Even at an extremely strict threshold of 0.90, which splits the RWS into an average of 15 hyper-fine-grained topics, the proportion of cross-topic cited papers is only 13.12\%. This result shows that in human RWS writing, over 86--91\% of cited papers do indeed belong to a single topic.

\section{LLM Prompts}
\label{app:prompts}
\subsection{Prompt for Graph of Thoughts Construction}

\begin{promptbox}[title={System Prompt: Graph-of-Thought Construction}]
You are an expert scientific assistant.

\textbf{\#\#\# Task}
You are constructing a Graph-of-Thought (GoT) for a given topic. Each paper provides a chain-of-thought (CoT) – an ordered list of steps. Build a Directed Acyclic Graph that follows these rules:

\begin{enumerate}
    \item There are two types of nodes: original and consensus.
    \begin{itemize}
        \item 1.1) Each Original node is one step in each CoT.
        \item 1.2) Each Consensus node is an aggregation of two or more similar original nodes.
    \end{itemize}
    
    \item There are two types of edges: sequential and consensus.
    \begin{itemize}
        \item 2.1) A sequential edge is an edge connecting two original nodes.
        \item 2.2) A consensus edge is an edge connecting an original node to a consensus node or a consensus node to another consensus node.
    \end{itemize}
    
    \item You should name all nodes starting from n0. The order of nodes in each CoT is determined.
    
    \item The content of a consensus node should be a summary of the content of the original nodes that comprise it. \textbf{IMPORTANT: The consensus node text MUST start with metadata from ALL contributing papers, formatted as: |||paper\_id:X; author:A; year:Y||| |||paper\_id:Z; author:B; year:W||| followed by the summary.} This preserves which papers contributed to the consensus.
    
    \item Once several original nodes are aggregated into a consensus node, the edges connecting the original nodes will be converted to connect the consensus node, regardless of whether they are outgoing or incoming edges, and the aggregated original nodes will be removed from the graph. The 'paper\_id' of the edge will not change; it stays the same as it is linked to the original nodes.
    
    \item The original nodes that make up the consensus nodes must come from different CoTs and cannot be different nodes under the same CoT.
    
    \item In each CoT, the first node will contain some metadata of the paper between "|||" markers. The metadata contains the paperid, author, and year. It helps you to understand where each node comes from because the graph is an acyclic directed graph and you can trace back to every node's predecessor to figure it out. Just keep it there and ignore it when doing content analysis.
    
    \item \textbf{IMPORTANT: Actively look for opportunities to create consensus nodes!} You should be AGGRESSIVE in merging similar steps across papers. Consider the following as candidates for consensus:
    \begin{itemize}
        \item Steps that discuss similar methodology or technique (even if worded differently)
        \item Steps that address a similar problem or challenge
        \item Steps that describe similar experimental setups or datasets
        \item Steps that present similar conclusions or findings
        \item Steps that reference similar background concepts
    \end{itemize}
    Do NOT require exact wording match - semantic similarity is sufficient. If two steps from different papers are talking about roughly the same idea, merge them into a consensus node. You should lower your threshold to merge nodes when there is not even one in each topic. If still can't find one, then you move on.
    
    \item In each original node there might be some section names indicating where does this node's idea come from, when merging, you can do whatever you want to keep or drop or change some form of the information to finally better maintain necessary information.
    
    \item Only return JSON following the given schema. No prose, no markdown.
    
    \item \textbf{Creating consensus nodes is a KEY objective.} A good GoT should have at least a few consensus nodes that capture shared ideas across papers. If you find no similar steps at all, double-check - papers on the same topic usually share some common ground (e.g., problem definition, evaluation metrics, baseline methods).
\end{enumerate}

You can refer to the following example to check the graph structure format:

\begin{quote}
\textbf{Example (two papers):}\\
Paper A: [A1, A2, A3] $\rightarrow$ nodes A\_step1, A\_step2, A\_step3\\
Paper B: [B1, B2, B3] $\rightarrow$ nodes B\_step1, B\_step2, B\_step3\\
Suppose A\_step2 $\approx$ B\_step3. Create n\_con1 and replace A\_step2, B\_step3 with n\_con1.\\
Edges should be:\\
A\_s1 $\rightarrow$ n\_con1 $\rightarrow$ A\_s3   and   B\_s1 $\rightarrow$ B\_step2 $\rightarrow$ n\_con1\\
(The above example's name of node is not what really in the graph, you should follow the requirement to name the nodes.)
\end{quote}

\textbf{\#\#\# Output format}
\begin{lstlisting}[language=json]
{
  "topic": "<topic string>",
  "nodes": [
    {"id": "n0", "text": "step content"},
    {"id": "n1", "text": "..."}
  ],
  "edges": [
    {"source": "n0", "target": "n1", "paper_id": "P001"},
    {"source": "n1", "target": "n2", "paper_id": "P001"}
  ]
}
\end{lstlisting}
\end{promptbox}

\subsection{Prompts for ACPN Construction}
\subsubsection{Claim Extraction Prompt}
\begin{promptbox}[title={System Prompt: Claim Extraction}]
You are a helpful research assistant designed to output JSON.

\textbf{\#\#\# Task}
Your task is to carefully read the provided text from a scientific paper and extract the main claims or hypotheses made by the authors. The paper is given in the format of \texttt{\{section name: section texts\}}; you should think of the paper as a whole, not section by section.

A \textbf{claim} is a statement of finding, discovery, or a core argument the paper is making. Extract between \textbf{3 to 7} of the most important claims.

\textbf{\#\#\# Constraints}
\begin{itemize}
    \item Paper text begins with metadata between "|||" markers - \textbf{ignore this metadata during analysis}.
    \item The paper sections are prioritized by importance (abstract, introduction, conclusion, results, methods) to ensure key claims are captured.
    \item Some papers may have \texttt{[...TRUNCATED...]} markers indicating omitted content to fit context limits.
    \item Your output MUST be a valid JSON object that strictly adheres to the schema provided.
    \item Do not add any other explanatory text, introductions, or markdown formatting like \texttt{```json}.
\end{itemize}

\textbf{\#\#\# Output format}
\begin{lstlisting}[language=json]
{
  "claims": [
    "Claim 1 text...", 
    "Claim 2 text..."
  ] 
}
\end{lstlisting}
\end{promptbox}

\subsubsection{Relation Prediction Prompt}
\begin{promptbox}[title={System Prompt: Relation Prediction}]
You are an expert reasoning engine. Your task is to decide the \textbf{direction-agnostic relation} between two scientific papers.

\textbf{\#\#\# Possible Labels}
\begin{itemize}
    \item \textbf{support} – the two papers make compatible or mutually reinforcing claims.
    \item \textbf{contrast} – one paper challenges, contradicts, or weakens the other's claims.
    \item \textbf{neutral} – neither clear support nor clear contrast.
\end{itemize}

The optional field \texttt{"additional\_context"} contains up to 1000 characters of Graph-of-Thought steps for each paper; treat it as high-level reasoning evidence. There might be metadata between "|||" markers - ignore this metadata during analysis.

Return \textbf{only} the strict JSON object shown in the output format. Do not add markdown, comments, or extra keys.

\textbf{\#\#\# Output format}
\begin{lstlisting}[language=json]
{ 
  "relation": "support" 
}
\end{lstlisting}
\end{promptbox}

\subsection{Prompt for RWS Generation}
\begin{promptbox}[title={System Prompt: Related Work Generator}]
You are an academic writing assistant. Your task is to write a well-structured Related Work section for the given TOPIC.

You have access to three kinds of information:
\begin{itemize}
    \item \textbf{Graph-of-Thought (GoT)} — a reasoning graph that organizes cited works into several topic-specific subgraphs. Identify the distinct topics in GoT. For each topic, determine which reference papers belong to it. Use GoT's logical structure to order them (precursor $\rightarrow$ complementary $\rightarrow$ contrasting).
    \item \textbf{Argument-Counter-argument Planning Network (ACPN)} — a directed graph connecting all references, where edges labeled \textit{support}, \textit{contrast}, or \textit{neutral} indicate argument relations across papers. Integrate this information to form comparative statements.
    \item \textbf{Source text excerpts} (abstract, introduction, conclusion, etc.) from the current paper. Use them to infer what the current work does and to highlight its novel contribution in contrast to prior work.
\end{itemize}

\vspace{0.5em}
\textbf{\#\#\# Citation style rules (CRITICAL - you MUST follow these exactly):}

The \texttt{'citations information'} field contains ALL valid references. Each reference has:
\begin{itemize}
    \item \texttt{'k'}: numeric citation key (integer like 1, 2, 3...)
    \item \texttt{'authors'}: list of author names (e.g., ['Trevor Cohn', 'Chris Dyer', ...])
    \item \texttt{'year'}: publication year (e.g., 2016)
    \item \texttt{'title'}: paper title
\end{itemize}

Determine citation style by checking the source text:
\begin{itemize}
    \item If \textbf{NUMERIC style} $\rightarrow$ use [k] format, e.g., [1], [2, 3], [1-4]. The 'k' values come from \texttt{'citations information'}.
    \item If \textbf{AUTHOR-YEAR style} $\rightarrow$ use (FirstAuthorSurname et al., year) or (FirstAuthorSurname, year) for single-author papers. Extract the FIRST author's SURNAME from the \texttt{'authors'} list in \texttt{'citations information'}.\\
    \textit{Example:} if authors=['Trevor Cohn', 'Chris Dyer'] and year=2016, cite as (Cohn et al., 2016).
    \item For multiple citations: [1, 2] or (Cohn et al., 2016; Tang et al., 2016).
\end{itemize}

\textbf{CRITICAL: You MUST use ONLY the author names and years that appear in \texttt{'citations information'}. If you cannot find a reference in \texttt{'citations information'}, do not cite it.}

\vspace{0.5em}
\textbf{\#\#\# Paragraph structure rules:}
\begin{itemize}
    \item Let the GoT topology guide your paragraph structure naturally.
    \item If GoT contains multiple DISCONNECTED subgraphs (separate topic clusters), write one paragraph per subgraph.
    \item If GoT is a single CONNECTED component, write as a cohesive single paragraph or use natural topic transitions.
    \item Do NOT artificially split or merge paragraphs. Let the logical structure of the content determine breaks.
\end{itemize}

\vspace{0.5em}
\textbf{\#\#\# Writing requirements:}
\begin{itemize}
    \item Write in formal academic tone and follow the structure conventions found in the source paper.
    \item Integrate GoT and ACPN insights coherently rather than list them.
    \item Highlight how the current paper differs from and improves upon prior work.
    \item Avoid redundancy and ensure the text reads naturally.
    \item Output length should be proportional to the graph content: write approximately 1-2 sentences per unique thought node in GoT. Do not pad or elaborate beyond what the graph provides. If the graph contains fewer nodes, produce correspondingly shorter output.
\end{itemize}

\vspace{0.5em}
\textbf{\#\#\# Three-step generation plan:}
\begin{enumerate}
    \item \textbf{Step 1:} Generate a comprehensive related work section using all available information.
    \item \textbf{Step 2:} Generate a more concise version that summarizes and compresses the first.
    \item \textbf{Step 3:} Compare the two versions and synthesize a final version that is brief yet complete, ensuring that no important content is lost.
    \begin{itemize}
        \item You should balance the sentence distribution to make it close to human-written text.
        \item Not only summarize prior work, but highlight the current paper's differences and transit smoothly between topics.
        \item In the final version, remove duplicated citation markers and keep only the first occurrence, as long as doing so does not reduce readability or create ambiguity about which paper is being referenced.
    \end{itemize}
\end{enumerate}
In each step, respect its description in the provided schema as a behavioral instruction.
\end{promptbox}

\section{Betweeness Centrality Threshold Tuning}
\label{app:centrality}

We tune the betweenness centrality threshold, varying it from 10\% to 50\%. While a stricter 10\% threshold yields marginally higher surface-level string matching metrics (ROUGE-1 F1 of 0.782 vs. 0.771 at 20\%), it starves the Steiner tree of necessary connecting nodes, causing declines in relationship recognition (citation intent macro F1 of 0.897). It also increases the ratio difference of Single Summary sentences (+1.25\%). 

However, increasing the centrality threshold above 20\% is even worse. Noisy, irrelevant nodes are retained in the graph, degrading the citation intent macro F1 to 0.881 at 40\% threshold (from 0.974 at 20\%) and increasing the ratio difference of Single Summaries to +1.56\% (from 0.011 at 20\%).

\section{Citation Post-Processing}
\label{app:citation-management}
Ensuring bibliographic accuracy is paramount. Our approach decouples citation \textit{placement} from citation \textit{formatting}. During generation, our Writer LLM is instructed to strictly adhere to the citation style found in the source text context. If the source utilizes numeric markers (e.g. ``[1]" ), the model generates numeric citations corresponding to the provided cited paper indices; if the source employs an author-year format, the model attempts to generate citations in that style directly. This adaptability minimizes generation friction by allowing the model to mimic the citation placement used in the rest of the target citing paper.

For each of these two citation formats, we apply a deterministic post-processing step:

\paragraph{Numeric-to-Text Conversion} If the generated text uses the numeric format, we map each numeric index back to the set of cited papers and substitute the numeric marker with the corresponding author-year string (e.g., ``(Smith et al., 2025)''). We also apply this post-processing step to the target RWS and all comparison systems' outputs to standardize the citation markers for evaluation.

\paragraph{Citation Verification} If the generated text uses the author-year format, we perform a validation check against the set of cited papers. We verify that the author-year pair exists in the input bibliography, flagging and correcting any hallucinations where the model may have invented non-existent dates or authors.


\section{Computational Cost}
\label{app:cost}

Our total inference time for the OARelatedWork test set is about 3 hours; as we run 16 samples concurrently, this averages out to 1.5 minutes per sample if run sequentially. The cost is around \$45 USD for 1878 examples, or 2.5 cents per sample, using gpt-4o-mini.

\section{Generation and Citation Importance Statistics}

\begin{table}[h]
  \centering
  \small
  \resizebox{\linewidth}{!}{%
    \begin{tabular}{lrrrrr}
    \toprule
    \textbf{Metric} & \textbf{Target} & \textbf{SRW} & \textbf{L\&O} & \textbf{GRASP (unpr.)} & \textbf{GRASP (pru.)} \\
    \midrule
    \# Sentences & 24,753 & 21,480 & 40,636 & 39,788 & 36,237 \\
    \midrule
    \# Dominant & 255   & 159 & 503 & 880 & 423 \\
    \hspace{5pt} \# Sent. w/ Dom. & 254   & 1058  & 503   & 849   & 423 \\
    \midrule
    \# Reference & 594   & 459 & 884 & 940 & 796 \\
    \hspace{5pt}\# Sent. w/ Ref. & 479   & 1356  & 742   & 757   & 643 \\
    \midrule
    \# Ref / \# Dom & 2.33  & 2.87  & 1.76  & 1.07  & 1.88 \\
    \bottomrule
    \end{tabular}%
  }
  \caption{Total RWS sentence and citation count statistics on the OARelatedWork test set.}
  \label{tab:generationstats}
\end{table}

Table \ref{tab:generationstats} shows that, in terms of total RWS length, the SRW baseline is extremely concise, generating only 87\% of the target sentence count. In contrast, both GRASP and especially L\&O are more verbose. The excessive length and disproportionately high number of Dominant citation spans generated by both L\&O and unpruned GRASP indicate a tendency to over-generate descriptions of individual cited papers, rather than focusing on inter-paper relationships. Pruned GRASP effectively reduces verbosity by filtering topologically peripheral GoT nodes, thereby reducing the sentence count and increasing the ratio of \textit{Reference} to \textit{Dominant} citations to 1.88, the closest to the target ratio. 

By comparing the number of \textit{Dominant}/\textit{Reference} citations to the number of sentences containing such citations, we can see that the target, L\&O, and GRASP all follow the observations of \citet{li-etal-2022-corwa}: most \textit{Dominant} citations occupy an entire sentence, while most \textit{Reference} citations cover less than a single sentence. Interestingly, we also see that SRW contains a very high ratio of citation sentences to citations; despite generating fewer total citations than the target, each citation spans multiple sentences, suggesting that cited papers are discussed with an unnecessary amounts of detail.




\section{Ablation Study Results}
\label{app:ablation}

These tables compare the performance of the pruned GRASP full model against ablated variants. Note that the \textit{GoT only} variant is pruned, and the Direct ACPN variant includes a GoT.

\begin{table}[h]
  \centering  
  \resizebox{0.98\linewidth}{!}{%
    \begin{tabular}{l cccc}
    \toprule
    \textbf{Metric} & \textbf{Full Model} & \textbf{ACPN only} & \textbf{GoT only} & \textbf{Direct ACPN} \\
    \midrule
    ROUGE-1 (P) & \textbf{0.653} & 0.441 & 0.465 & 0.627 \\
    ROUGE-1 (R) & 0.978 & 0.724 & \textbf{0.983} & 0.955 \\
    ROUGE-1 (F1) & \textbf{0.771} & 0.515 & 0.619 & 0.757 \\
    \addlinespace
    ROUGE-2 (P) & 0.644 & 0.298 & 0.456 & 0.642 \\
    ROUGE-2 (R) & 0.974 & 0.549 & \textbf{0.975} & 0.938 \\
    ROUGE-2 (F1) & \textbf{0.766} & 0.365 & 0.610 & 0.762 \\
    \addlinespace
    ROUGE-L (P) & \textbf{0.648} & 0.338 & 0.459 & 0.571 \\
    ROUGE-L (R) & 0.975 & 0.600 & \textbf{0.977} & 0.826 \\
    ROUGE-L (F1) & \textbf{0.769} & 0.407 & 0.613 & 0.675 \\
    \addlinespace
    BERTScore (P)   & \textbf{0.738} & 0.356 & 0.702 & 0.718 \\
    BERTScore (R)   & \textbf{0.932} & 0.495 & 0.919 & 0.920 \\
    BERTScore (F1)   & \textbf{0.832} & 0.423 & 0.806 & 0.816 \\
    \addlinespace
    BLEU (Macro)     & \textbf{63.42} & 29.89 & 46.61 & 56.35 \\
    BLEU (Micro)     & \textbf{68.09} & 33.91 & 50.21 & 63.21 \\
    \addlinespace
    METEOR (Macro)   & \textbf{0.778} & 0.468 & 0.670 & 0.686 \\
    METEOR (Micro)   & \textbf{0.509} & 0.427 & 0.475 & 0.473 \\
    \bottomrule
    \end{tabular}%
  }
    \caption{Ablation study results using traditional text generation metrics.}
  \label{tab:ablation-text}
\end{table}

In Table \ref{tab:ablation-text}, we see that the GoT-only variant maintains comparable performance on ROUGE and BERTScore recall but suffers from lowered precision, while the ACPN-only variant has significantly lower precision and recall. The GoT encodes the cited paper content, so the GoT-only model is able to generate most of the target RWS content. In contrast, the ACPN provides higher-level guidance on inter-paper argumentative relationships, guiding the full model to focus on the most salient aspects of the cited papers, but the ACPN alone does not capture enough cited paper content to generate a good RWS.

\begin{table}[h]
  \centering  
  \resizebox{0.98\linewidth}{!}{%
    \begin{tabular}{l cccc}
    \toprule
    \textbf{Metric} & \textbf{Full Model} & \textbf{ACPN only} & \textbf{GoT only} & \textbf{Direct ACPN} \\
    \midrule
    Single\_summ    & \textbf{0.011} & 0.064 & 0.050 & -0.011  \\
    Multi\_summ     & \textbf{0.002} & 0.003 & \textbf{0.002} & 0.003 \\
    Reflection      & \textbf{-0.003}& 0.013 & 0.023 & 0.015 \\
    Narrative\_cite & \textbf{0.000} & -0.013 & \textbf{0.000} & 0.001 \\
    Transition      & \textbf{0.002} & -0.014& 0.020 & 0.037 \\
    \bottomrule
    \end{tabular}%
  }
    \caption{Ablation study results comparing sentence discourse roles. Scores closer to 0 are better for ratio difference.}
  \label{tab:ablation-discourse}
\end{table}

In Table \ref{tab:ablation-discourse}, we see that both variants over-generate single cited paper summaries, suggesting that both graph layers are needed to focus the Writer module on inter-paper relationships. Meanwhile, the Direct ACPN variant over-produces \textit{Transition}- and \textit{Reflection}-type filler sentences because it lacks concrete extracted claims to ground generation.

\begin{table}[h]
  \centering  
  \resizebox{0.98\linewidth}{!}{%
    \begin{tabular}{l cccc}
    \toprule
    \textbf{Metric} & \textbf{Full Model} & \textbf{ACPN only} & \textbf{GoT only} & \textbf{Direct ACPN} \\
    \midrule
    Dominant (P)      & \textbf{0.917} & 0.776 & 0.827 & 0.812 \\
    Dominant (R)      & \textbf{0.928} & 0.704 & 0.749 & 0.877 \\
    Dominant (F1)     & \textbf{0.922} & 0.738 & 0.817 & 0.843 \\
    \addlinespace
    Reference (P)     & \textbf{0.917} & 0.684 & 0.717 & 0.720 \\
    Reference (R)     & \textbf{0.983} & 0.838 & 0.746 & 0.914 \\
    Reference (F1)    & \textbf{0.949} & 0.753 & 0.731 & 0.805 \\
    \bottomrule
    \end{tabular}%
  }
    \caption{Ablation study results for citation importance fidelity.}
  \label{tab:ablation-domref}
\end{table}

In Table \ref{tab:ablation-domref}, we see that the ACPN-only model performs slightly better on \textit{Reference}-type citations, while GoT-only performs slightly better on \textit{Dominant} citations, again showing that the GoT provides most of the cited paper content, while the ACPN provides higher-level inter-paper relationships.

\begin{table}[h]
  \centering  
  \resizebox{0.98\linewidth}{!}{%
    \begin{tabular}{l cccc }
    \toprule
    \textbf{Metric} & \textbf{Full Model} & \textbf{ACPN only} & \textbf{GoT only} & \textbf{Direct ACPN} \\
    \midrule
    Background (P)    & \textbf{0.728} & 0.302 & 0.714 & 0.562  \\
    Background (R)    & 0.964 & 0.519 & \textbf{0.965} & 0.961 \\
    Background (F1)   & \textbf{0.829} & 0.382 & 0.821 & 0.709 \\
    \addlinespace
    Differences (P)   & \textbf{0.844} & 0.203 & 0.748 & 0.390 \\
    Differences (R)   & \textbf{0.960} & 0.464 & \textbf{0.960} & 0.944 \\
    Differences (F1)  & \textbf{0.898} & 0.283 & 0.841 & 0.710 \\
    \addlinespace
    Extends (P)       & \textbf{0.862} & 0.283 & 0.841 & 0.177 \\
    Extends (R)       & 0.980 & 0.700 & \textbf{1.000} & 1.000 \\
    Extends (F1)      & \textbf{0.917} & 0.400 & 0.836 & 0.301 \\
    \addlinespace
    Future Work (P)   & \textbf{1.000} & 0.000 & 0.750 & 0.500 \\
    Future Work (R)   & \textbf{1.000} & 0.000 & \textbf{1.000} & 1.000 \\
    Future Work (F1)  & \textbf{1.000} & 0.000 & 0.857 & 0.666 \\
    \addlinespace
    Motivation (P)    & 0.684 & 0.455 & \textbf{0.909} & 0.363 \\
    Motivation (R)    & 0.983 & 0.581 & \textbf{0.987} & 0.978 \\
    Motivation (F1)   & 0.806 & 0.510 & \textbf{0.947} & 0.529 \\
    \addlinespace
    Similarities (P)  & \textbf{0.955} & 0.680 & 0.711 & 0.894 \\
    Similarities (R)  & 0.955 & 0.674 & \textbf{0.959} & 0.948 \\
    Similarities (F1) & \textbf{0.955} & 0.103 & 0.817 & 0.920 \\
    \addlinespace
    Uses (P)          & \textbf{0.991} & 0.821 & 0.958 & 0.616 \\
    Uses (R)          & \textbf{0.950} & 0.319 & 0.949 & 0.932 \\
    Uses (F1)         & \textbf{0.970} & 0.460 & 0.954 & 0.741 \\
    \bottomrule
    \end{tabular}%
  }
    \caption{Ablation study results for citation intent fidelity.}
  \label{tab:ablation-intent}
\end{table}

In Table \ref{tab:ablation-intent}, we see that the GoT-only variant performs comparably to the full model (or even slightly better, in the cast of \textit{Motivation} citations), while the ACPN-only variant is much worse. Again we conclude that, since the ACPN captures argumentative relationships between cited papers, but very little paper content, the quality of individual citations generated by the ACPN-only variant is relatively poor. Meanwhile, the Direct ACPN variant’s ability to match ground truth citation intent collapses without explicit claims to anchor comparisons among cited papers, with especially severe degradation relations that require specific evidence, such as \textit{Similarities/Differences} and \textit{Extends}.

\begin{table}[h]
  \centering  
  \resizebox{0.98\linewidth}{!}{%
    \begin{tabular}{l cccc}
    \toprule
    \textbf{Metric} & \textbf{Full Model} & \textbf{ACPN only} & \textbf{GoT only} & \textbf{Direct ACPN} \\
    \midrule
    Edge Jaccard    & \textbf{0.847} & 0.665 & 0.689 & 0.552 \\
    Kendall's $\tau$& \textbf{0.886} & 0.662 & 0.666 & 0.710 \\
    \bottomrule
    \end{tabular}%
  }
    \caption{Ablation study results for citation grouping and ordering.}
  \label{tab:ablation-cooccurrence}
\end{table}

In Table \ref{tab:ablation-cooccurrence}, we see that both variants perform similarly in terms of citation grouping and ordering, with GoT-only pulling slightly ahead. Both graphs contain grouping and ordering information via edges between cited papers/thoughts and complement each other to produce the strong performance of the full model.

\section{Comparison of Compressed and Final Drafts}
\begin{table}[t] 
\centering
  \footnotesize  
    \begin{tabular}{l cc}
    \toprule
    \textbf{Metric} & \textbf{Compressed} & \textbf{Final} \\
    \midrule
    ROUGE-1 (P) & \textbf{0.684} & 0.653 \\
    ROUGE-1 (R) & 0.977 & \textbf{0.978} \\
    ROUGE-1 (F1)& \textbf{0.794} & 0.771 \\
    \addlinespace
    ROUGE-2 (P) & \textbf{0.676} & 0.644 \\
    ROUGE-2 (R) & 0.973 & \textbf{0.974} \\
    ROUGE-2 (F1)& \textbf{0.789} & 0.766 \\
    \addlinespace
    ROUGE-L (P) & \textbf{0.679} & 0.648 \\
    ROUGE-L (R) & \textbf{0.975} & \textbf{0.975} \\
    ROUGE-L (F1)& \textbf{0.791} & 0.769 \\
    \addlinespace
    BERTScore (P)   & \textbf{0.750} & 0.738 \\
    BERTScore (R)   & \textbf{0.933} & 0.932 \\
    BERTScore (F1)  & \textbf{0.838} & 0.832 \\
    \addlinespace
    BLEU (Macro)    & \textbf{66.12} & 63.42 \\
    BLEU (Micro)    & \textbf{71.01} & 68.09 \\
    METEOR (Macro)  & \textbf{0.795} & 0.778 \\
    METEOR (Micro)  & \textbf{0.514} & 0.509 \\
    \bottomrule
    \end{tabular}%
    \caption{Comparison of compressed and final drafts using traditional text generation metrics.}
  \label{tab:draft-text}
\end{table}

\begin{table}[t] 
\centering
  \footnotesize  
    \begin{tabular}{l cc}
    \toprule
    \textbf{Metric} & \textbf{Compressed} & \textbf{Final} \\    
    \midrule
    Single\_summ    & \textbf{0.011} & \textbf{0.011} \\
    Multi\_summ     & 0.003 & \textbf{0.002} \\
    Reflection      & -0.003 & -0.003 \\
    Narrative\_cite & 0.001 & \textbf{0.000} \\
    Transition      & 0.002 & 0.002 \\
    \bottomrule
    \end{tabular}%
    \caption{Comparison of compressed and final drafts on sentence discourse roles. Scores closer to 0 are better for ratio difference.}
  \label{tab:darft-discourse}
\end{table}

\begin{table}[t] 
\centering
  \footnotesize  
    \begin{tabular}{l cc}
    \toprule
    \textbf{Metric} & \textbf{Compressed} & \textbf{Final} \\    
    \midrule
    Dominant (P)     & \textbf{0.951} & 0.917 \\
    Dominant (R)     & \textbf{0.944} & 0.928 \\
    Dominant (F1)    & \textbf{0.951} & 0.922 \\
    \addlinespace
    Reference (P)    & \textbf{0.981} & 0.917 \\
    Reference (R)    & 0.945 & \textbf{0.983} \\
    Reference (F1)   & \textbf{0.963} & 0.949 \\
    \bottomrule
    \end{tabular}%
    \caption{Comparison of compressed and final drafts on citation importance fidelity.}
  \label{tab:draft-domref}
\end{table}

\begin{table}[t] 
\centering
  \footnotesize  
    \begin{tabular}{l cc}
    \toprule
    \textbf{Metric} & \textbf{Compressed} & \textbf{Final} \\    
    \midrule
    Background (P)   & 0.725 & \textbf{0.728} \\
    Background (R)   & 0.962 & \textbf{0.964} \\
    Background (F1)  & 0.827 & \textbf{0.829} \\
    \addlinespace
    Differences (P)  & 0.841 & \textbf{0.844} \\
    Differences (R)  & 0.960 & 0.960 \\
    Differences (F1) & 0.890 & \textbf{0.898} \\
    \addlinespace
    Extends (P)      & 0.823 & \textbf{0.862} \\
    Extends (R)      & \textbf{1.000} & 0.980 \\
    Extends (F1)     & 0.903 & \textbf{0.917} \\
    \addlinespace
    Future Work (P)  & 0.750 & \textbf{1.000} \\
    Future Work (R)  & 1.000 & 1.000 \\
    Future Work (F1) & 0.857 & \textbf{1.000} \\
    \addlinespace
    Motivation (P)   & \textbf{0.690} & 0.684 \\
    Motivation (R)   & 0.983 & 0.983 \\
    Motivation (F1)  & \textbf{0.811} & 0.806 \\
    \addlinespace
    Similarities (P) & \textbf{0.959} & 0.955 \\
    Similarities (R) & 0.955 & 0.955 \\
    Similarities (F1)& \textbf{0.957} & 0.955 \\
    \addlinespace
    Uses (P)         & 0.982 & \textbf{0.991} \\
    Uses (R)         & 0.950 & 0.950 \\
    Uses (F1)        & 0.965 & \textbf{0.970} \\
    \bottomrule
    \end{tabular}%
    \caption{Comparison of compressed and final drafts on citation intent fidelity.}
  \label{tab:draft-intent}
\end{table}

\begin{table}[t] 
\centering
  \footnotesize  
    \begin{tabular}{l cc}
    \toprule
    \textbf{Metric} & \textbf{Compressed} & \textbf{Final} \\    
    \midrule
    Edge Jaccard     & 0.843 & \textbf{0.847} \\
    Kendall's $\tau$ & 0.866 & \textbf{0.886} \\
    \bottomrule
    \end{tabular}%
    \caption{Comparison of compressed and final drafts on citation grouping and ordering.}
  \label{tab:draft-cooccurr}
\end{table}

\label{app:drafts}
In Tables \ref{tab:draft-text}-\ref{tab:draft-cooccurr}, we see only small differences in metric performance between the Compressed and Final drafts produced by our Writer module. Specifically, the Final draft trades a slight dip in performance on text generation metrics for slightly improved citation intent fidelity, discourse ratios, and citation grouping/ordering. 

Figure \ref{fig:example-compressed} shows the Compressed draft of the example RWS. We can see that the it uses some ambiguous or unexpected phrasing in order to reduce the overall word count, often falling into noticeably AI-like word choice, compared to the more expressive Final draft shown in Figure \ref{fig:example-pruned}.

\section{Example Related Work Sections}
\label{app:generated-examples}

These example RWS are generated for ``Parameter Sharing Methods for Multilingual Self-Attentional Translation Models" \cite{sachan-neubig-2018-parameter}. This example also produced the GoT excerpt shown in Figure \ref{fig:consensus}; it was chosen to be easily understandable by an NLP audience and may not demonstrate every distinction among approaches discussed in Section \ref{sec:results}.

Note that the target RWS (Figure \ref{fig:example-target}) has three sentences removed, due to their containing citations whose cited papers are not available in OARelatedWork.

\FloatBarrier

\begin{figure*}[t]
    \centering
    \fbox{%
        \begin{minipage}{\dimexpr\textwidth-2\fboxsep-2\fboxrule}
            In this section, we will review the prior work related to MTL and multilingual translation.\\

Ando and Zhang (2005) obtained excellent results by adopting an MTL framework to jointly train linear models for NER, POS tagging, and language modeling tasks involving some degree of parameter sharing. Later, Collobert et al. (2011) applied MTL strategies to neural networks for tasks such as POS tagging, NER, and chunking by sharing the sequence encoder and reported moderate improvements in results. [Sentences removed due to cited papers missing from OARelatedWork.] MTL has also been widely applied to multilingual translation that will be discussed next.\\

On the multilingual translation task, Dong et al. (2015) obtained significant performance gains by sharing the encoder parameters of the source language while having a separate decoder for each target language. Later, Firat et al. (2016) attempted the more challenging task of many-to-many translation by training a model that consisted of one shared encoder and decoder per language and a shared attention layer that was common to all languages. This approach obtained competitive BLEU scores on ten European language pairs while substantially reducing the total parameters. Recently, Johnson et al. (2017) proposed a unified model with full parameter sharing and obtained comparable or better performance compared with bilingual translation scores. During model training and decoding, target language was specified by an additional token at the beginning of the source sentence. Coming to low-resource language translation, Zoph et al. (2016) used a transfer learning approach of fine-tuning the model parameters learned on a high-resource language pair of French→English and were able to significantly increase the translation performance on Turkish and Urdu languages. Recently, Gu et al. (2018) addresses the many-to-one translation problem for extremely low-resource languages by using a transfer learning approach such that all language pairs share the lexical and sentence-level representations. By performing joint training of the model with high-resource languages, large gains in the BLEU scores were reported for low-resource languages. \\

In this paper, we first experiment with the Transformer model for one-to-many multilingual translation on a variety of language pairs and demonstrate that the approach of Johnson et al. (2017) and Dong et al. (2015) is not optimal for all kinds of target-side languages. Motivated by this, we introduce various parameter sharing strategies that strike a happy medium between full sharing and partial sharing and show that it achieves the best translation accuracy.
        \end{minipage}%
    }
    \caption{Cleaned target related work section from ``Parameter Sharing Methods for Multilingual Self-Attentional Translation Models" \cite{sachan-neubig-2018-parameter}.}
    \label{fig:example-target}
\end{figure*}

\begin{figure*}[t]
    \centering
    \fbox{%
        \begin{minipage}{\dimexpr\textwidth-2\fboxsep-2\fboxrule}
            Recent advancements in neural machine translation (NMT) have prominently featured zero-resource and low-resource translation methods. Collobert et al. (2011) laid the groundwork for NMT by emphasizing the necessity of robust architectures to tackle diverse linguistic challenges. Firat et al. (2016) complemented this with a multi-way multilingual NMT model that employs a shared attention mechanism, improving efficiency and performance in zero-shot translation scenarios. Zoph et al. (2016) explored transfer learning strategies, highlighting how leveraging high-resource language pairs can considerably enhance BLEU scores for low-resource languages, effectively addressing data scarcity issues.\\
            
Further, Gu et al. (2018) reinforced the significance of shared representations in low-resource translations, aligning with the insights of Dong et al. (2015) regarding multi-task learning frameworks that optimize performance through the use of unlabeled data. Johnson et al. (2016) also advocated for semi-supervised learning techniques to boost model performance, echoing Zoph et al.'s findings. In contrast to these established approaches, our paper introduces a novel parameter sharing strategy within the Transformer architecture, strategically focusing on specific parameters to enhance multilingual translation accuracy across both closely and distantly related languages.
        \end{minipage}%
    }
    \caption{Related work section generated by pruned GRASP.}
    \label{fig:example-pruned}
\end{figure*}

\begin{figure*}[t]
    \centering
    \fbox{%
        \begin{minipage}{\dimexpr\textwidth-2\fboxsep-2\fboxrule}
            Recent advancements in neural machine translation (NMT) have prominently featured zero-resource and low-resource translation methods. Collobert et al. (2011) introduced foundational NMT techniques emphasizing robust architectures for diverse linguistic challenges. Firat et al. (2016) complemented this by proposing a multi-way multilingual NMT model with a shared attention mechanism, enhancing efficiency and competitive performance in zero-shot translation settings. Zoph et al. (2016) explored transfer learning strategies, demonstrating that high-resource language pairs can significantly boost BLEU scores for low-resource languages, thus addressing data scarcity. \\
            
            Gu et al. (2018) further highlighted the importance of shared representations in low-resource translations, aligning with Dong et al. (2015) on multi-task learning frameworks that optimize performance using unlabeled data. Moreover, Johnson et al. (2016) indicated the potential of semi-supervised learning techniques to improve model performance, complementing the findings of Zoph et al. (2016). Contrasting with these approaches, our work proposes a novel parameter sharing strategy within the Transformer architecture, focusing on sharing specific parameters to enhance multilingual translation accuracy across diverse language pairs.
        \end{minipage}%
    }
    \caption{Compressed draft related work section generated by pruned GRASP.}
    \label{fig:example-compressed}
\end{figure*}

\begin{figure*}[t]
    \centering
    \fbox{%
        \begin{minipage}{\dimexpr\textwidth-2\fboxsep-2\fboxrule}
            Natural Language Processing (NLP) has seen significant advancements in recent years, particularly in the realm of Neural Machine Translation (NMT). The foundational work by Collobert et al. (2011) established key principles of NMT, emphasizing the simplicity of implementation and end-to-end training capabilities. This work laid the groundwork for subsequent research that explored more advanced architectures and methodologies. \\

One of the pivotal developments in NMT was the introduction of the Transformer model by Vaswani et al. (2017), which revolutionized the field by replacing recurrent neural networks (RNNs) with self-attention mechanisms. This transition, as noted in the current work, has facilitated faster processing and improvements in translation accuracy across multilingual tasks. Specifically, the work of Firat et al. (2016) exemplified the application of a shared attention mechanism in multilingual settings, supporting the notion that joint training on multiple languages can enhance translation capabilities. \\

Building on these ideas, our research presents a novel framework that incorporates a shared encoder with multiple decoders, maintaining some shared parameters while allowing flexibility for distinct target languages. This approach is a significant extension of the multi-task learning paradigm previously explored by Dong et al. (2015), who utilized a shared encoder for multiple target languages, albeit with separate decoders. By contrast, our model introduces a hybrid structure that optimally balances shared and unique parameters, demonstrating superior performance in translating unrelated language pairs, a challenge highlighted in the works of Gu et al. (2018) and Johnson et al. (2016). \\

Furthermore, the concept of transfer learning for low-resource languages, as discussed by Zoph et al. (2016), aligns with our findings that partial sharing of parameters can be particularly beneficial when translating between languages from distant families. Our experimental results support this, showing that while full parameter sharing may yield satisfactory results for closely related languages, a more tailored approach is essential for achieving optimal accuracy in diverse linguistic contexts. \\

In summary, while prior research has extensively covered various architectures and training methodologies for NMT, the current work advances the understanding of parameter sharing strategies within the Transformer framework, thereby providing a more nuanced approach to multilingual translation. Our findings not only validate existing theories but also propose innovative solutions that enhance the efficacy of NMT systems in practical applications. This research contributes to the growing body of knowledge in NLP, particularly in the context of Arabic and multilingual texts, where effective translation strategies are paramount.
        \end{minipage}%
    }
    \caption{Related work section generated by unpruned GRASP.}
\end{figure*}

\begin{figure*}[t]
    \centering
    \fbox{%
        \begin{minipage}{\dimexpr\textwidth-2\fboxsep-2\fboxrule}
            In recent years, there has been substantial research into multilingual translation and the application of parameter sharing techniques. Early work by Ando and Zhang (2005) and Collobert et al. (2008) established the effectiveness of multi-task learning (MTL) frameworks, demonstrating that sharing parameters across tasks can enhance accuracy in various applications, including named entity recognition (NER) and part-of-speech (POS) tagging. These foundational studies laid the groundwork for subsequent investigations into multilingual neural machine translation (NMT). \\

Dong et al. (2015) explored parameter sharing within the context of multilingual translation, achieving performance gains by sharing encoder parameters while maintaining separate decoders for different languages. This approach was further advanced by Firat et al. (2016), who introduced a many-to-many translation model that shared encoders, decoders, and attention layers, resulting in competitive BLEU scores with a reduced parameter count. Building on this, Johnson et al. (2017) proposed a unified model that fully shared parameters across languages, demonstrating comparable performance to bilingual models and highlighting the potential of comprehensive parameter sharing in multilingual settings. \\

The challenge of low-resource languages also garnered attention, with Zoph et al. (2016) and Gu et al. (2018) employing transfer learning techniques to improve translation outcomes for languages with limited training data by leveraging resources from high-resource languages. Their work underscored the importance of parameter sharing in enhancing the robustness of translation systems in multilingual contexts. \\

Our study contributes to this growing body of work by examining various parameter sharing strategies specifically within self-attentional Transformer models. We investigate how different sharing approaches can optimize translation accuracy, particularly when dealing with languages from diverse families, thus addressing the limitations of full parameter sharing identified in earlier research. By integrating insights from previous studies, we aim to provide a balanced perspective on how to effectively leverage parameter sharing in multilingual translation tasks.
        \end{minipage}%
    }
    \caption{Related work section generated by the \citet{li-ouyang-2025-explaining} baseline.}
\end{figure*}

\begin{figure*}[t]
    \centering
    \fbox{%
        \begin{minipage}{\dimexpr\textwidth-2\fboxsep-2\fboxrule}
            The field of multilingual neural machine translation (NMT) has seen significant advancements through various architectural and training innovations aimed at improving performance across multiple language pairs. A key development in this domain is the proposal of a single neural translation model capable of handling multiple languages, as demonstrated by the work of Firat et al. (2016) and Johnson et al. (2016). Both studies utilize shared components, such as a unified attention mechanism and shared wordpiece vocabulary, to facilitate translation across diverse languages without altering the core architecture of the NMT system. These approaches stand in contrast to the more traditional encoder-decoder frameworks, which, as noted in Zoph et al. (2016), often fall short in low-resource scenarios. Here, transfer learning emerges as a crucial strategy, where parameters from high-resource language pairs are adapted to improve translation quality for low-resource pairs. \\

The challenge of translating a single source language into multiple target languages is addressed by Dong et al. (2015) through a multi-task learning model that leverages a shared encoder, illustrating the potential for improved translation quality via shared semantic and syntactic representations. This concept of parameter sharing resonates with the findings of Rie et al. (2005), which explore structural learning without empirical risk minimization, emphasizing the importance of structural parameters in enhancing predictive performance. \\

Further extending the paradigm of multilingual translation, Gu et al. (2018) introduces a transfer-learning approach that facilitates lexical and sentence-level sharing across languages with limited parallel data, underscoring the utility of universal representations in bridging resource gaps. This aligns with the broader trend towards reducing task-specific engineering, as advocated by Collobert et al. (2011), which promotes a unified network architecture across various natural language processing tasks, including translation. \\

Collectively, these studies highlight the efficacy of shared architectures and transfer learning in overcoming the limitations of multilingual NMT systems, demonstrating that strategic parameter sharing can significantly enhance translation performance across diverse language pairs.
        \end{minipage}%
    }
    \caption{Related work section generated by the Select, Read, Write \cite{liu-etal-2025-select} baseline.}
\end{figure*}

\end{document}